\documentclass[conference]{IEEEtran}
\IEEEoverridecommandlockouts
\usepackage{cite}
\usepackage{amsmath,amssymb,amsfonts}

\usepackage{graphicx}
\usepackage{subcaption}
\usepackage{makecell}
\usepackage{textcomp}
\usepackage{xcolor}

\usepackage{algorithm}

\usepackage{amssymb}
\usepackage{pifont}

\def\BibTeX{{\rm B\kern-.05em{\sc i\kern-.025em b}\kern-.08em
    T\kern-.1667em\lower.7ex\hbox{E}\kern-.125emX}}
\begin{document}

\title{An Ensembled Penalized Federated Learning Framework for Falling People Detection
}

\author{%
  Sizhe Rao,~Runqiu Zhang,~Sajal Saha,~\IEEEmembership{Senior Member,~IEEE},~and Liang Chen,~\IEEEmembership{Senior Member,~IEEE}\\[1ex]
  \IEEEauthorblockA{Department of Computer Science, University of Northern British Columbia, Prince George, BC, Canada}
}

\maketitle

\begin{abstract}

Falls among elderly and disabled individuals remain a leading cause of injury and mortality worldwide, necessitating robust, accurate, and privacy-aware fall detection systems. Traditional fall detection approaches, whether centralized or point-wise, often struggle with key challenges such as limited generalizability, data privacy concerns, and variability in individual movement behaviors. To address these limitations, we propose EPFL—an Ensembled Penalized Federated Learning framework that integrates continual learning, personalized modeling, and a novel Specialized Weighted Aggregation (SWA) strategy. EPFL leverages wearable sensor data to capture sequential motion patterns while preserving user privacy through homomorphic encryption and federated training. Unlike existing federated models, EPFL incorporates both penalized local training and ensemble-based inference to improve inter-client consistency and adaptability to behavioral differences. Extensive experiments on a benchmark fall detection dataset demonstrate the effectiveness of our approach, achieving a Recall of 88.31\% and an F1-score of 89.94\%, significantly outperforming both centralized and baseline models. This work presents a scalable, secure, and accurate solution for real-world fall detection in healthcare settings, with strong potential for continuous improvement via its adaptive feedback mechanism.
\end{abstract}


\section{Introduction}
Due to changes in traditional family structures, the number of older individuals living alone has significantly increased over the past few decades \cite{b1}. According to the report from World Health Organization (WHO) \cite{b2}, falls are the second leading cause of unintentional injury deaths worldwide, with particularly high morbidity among individuals aged 60 and older. Resulting in severe injuries, including fractures, head trauma, and even death, falls can significantly decline the quality of life of older adults \cite{b3}. Considering this, the need for effective monitoring and fall detection systems has been raised by this change aiming to ensure the safety of seniors. 


Falls can have long-term impacts on individuals, including significant disability-adjusted life years (DALYs) and high financial costs. According to the report\cite{b2}, falls cause over 38 million DALYs lost annually worldwide. In Canada, a 20\% reduction in falls could save approximately US\$120 million each year. Considering the severe injuries, potential fatalities and other additional costs resulting from sudden falls \cite{b4}, fall detection is a critical research area, especially for the elderly and individuals with disabilities. Effective fall detection methods are essential to reduce the severity and risk of falls. The reduction could essentially boost the quality of life for individuals by ensuring their safety and maintaining their functional independence. In practical circumstances, wearable devices (such as smart watches and fitness trackers) and ambient sensors placed in the living environment could help to monitor and detect sudden changes in movement and orientation. By utilizing the sequential position data from wearable devices and video/image data from ambient sensors, machine learning models can efficiently detect falls by treating the problem as time-series anomaly detection and human motion tracking respectively.

In this paper, we will mainly focus on sequential-based fall detection given the data collected from wearable devices. Currently, there are several rough challenges in fall detection:
\begin{itemize}
    \item Accuracy and Reliability: Achieving high accuracy remains a major challenge in fall detection research. In some scenarios, it is hard to differentiate between normal activities and actual falls, such as lying down quickly and sudden falls \cite{b5}. The mis-recognition could lead to high false-positive or false-negative rates, which undermines the reliability of the detection system.  Considering above, we would need to find the trade-off to balance the accuracy and reliability \cite{b6} to ensure the effectiveness of fall detection systems.
    
    \item Privacy Concerns: Fall detection systems often rely on various sensors, including wearable devices and ambient cameras, to monitor individual movements and detect falls. However, these systems raise significant privacy and data security concerns due to the collection, storage, and processing of sensitive personal data\cite{b7}. In addition, the risk of data breaches or unauthorized access to sensitive information further heightens these concerns \cite{b8}. Therefore, novel approaches are required to respect user privacy while ensuring the effectiveness of these systems.
    
    \item Individual Behaviour Differences: Variability in movement patterns, physical abilities, and daily routines among users can significantly affect the accuracy of fall detection. These behavioural differences can lead to reliability reduction in fall detection systems. To address this problem, personalized fall detection system \cite{b9} could probably help to more accuractely detect falls across diverse individuals to ensure higher reliability and effectiveness in real-world fall detection applications.
    
\end{itemize}

Many studies have tackled these issues. \cite{b24} applied Multi-step Histogram-based Outlier Scores to achieve 12.85\% Recall and 18.73\% F1-score while \cite{b38} employed Incremental Kolmogorov-Smirnov Test to reach around 50\% Accuracy. \cite{b36} attained 72\% Accuracy with machine learning agents. Using the same dataset we used as our project, these studies indicate room for optimization. \cite{b27,b28} addressed privacy concerns using federated learning; however, their approaches require client-side model training and fail to account for individual behavioral differences. To further address above problems, we propose a novel approach for falling people detection and the main contributions are stated below:

\begin{itemize}
    \item We propose EPFL—an Ensembled Penalized Federated Learning framework for fall detection, designed to address the challenges of privacy preservation, model robustness, and user-specific variability. EPFL integrates homomorphic encryption, penalized local training via FedProx, and ensemble-based inference to improve detection performance in real-world environments.
    
    \item We develop a novel Specialized Weighted Aggregation (SWA) strategy that combines trimmed mean filtering, training-epoch normalization, and exponential moving average fusion. This improves the robustness and stability of the federated learning process in the presence of noisy or non-IID client data.
    
    \item We introduce a dual-model ensemble approach that averages predictions from both global and client-specific models during inference. This personalization strategy significantly improves consistency and recall across heterogeneous clients without compromising privacy.

    \item EPFL incorporates a user feedback-driven continual learning loop, allowing both global and local models to evolve over time based on new user-verified events. This ensures adaptive and responsive performance in dynamic healthcare settings.

    \item Through extensive experiments on a publicly available fall detection dataset, our approach demonstrates superior performance (Recall = 88.31\%, F1-score = 89.94\%) compared to both classical models and existing federated learning baselines, validating its effectiveness and real-world applicability.

\end{itemize}
The remainder of this paper is structured as follows. Section II provides a literature review, discussing background knowledge related to fall detection systems and recent advancements using wearable sensors, ambient sensors, and federated learning techniques. Section III introduces the proposed Ensembled Penalized Federated Learning (EPFL) framework, including its architectural components, data preprocessing strategies, specialized aggregation method, and local inference process. Section IV describes the experimental setup, including dataset details, evaluation metrics, software and hardware environment, and presents a comprehensive performance analysis of both baseline and proposed models. Section V concludes the paper by summarizing the key findings, addressing limitations, and outlining potential directions for future work.

\section{Literature Review}

Fall detection systems can be divided into two main parts: the client part and the server part. The client part represents the user devices for real-time monitoring and alerting, such as wearable sensors and ambient sensors. The server part handles back-end tasks including data analysis, model training, model inference and continual model optimization\cite{b10}. In some scenarios, if the user devices are capable of offline inference, then model inference could be performed in the client part with dispatched weights from the server \cite{b11}. 

In the current literarture, research works based on different types of clients (user devices) would be given to provide a general view of fall detection research. Over the last decades, various types of user devices are invented that could be used to monitor the body movements, which can be categorised into two types of sensors: wearable sensors and ambient sensors \cite{b12}. 

Wearable sensors (such as fitness trackers, smart watches, connected headsets, smart glasses, wrist brands, etc) consist a good amount of sensors providing continuous data about the environmental variables such as body movements \cite{b13}. The movements are captured by the sequential position data in 3D space (x, y, z). Some research apply simple supervised machine learning classifiers such as Support Vector Machine (SVM) \cite{b14} and Random Forest \cite{b15} for single-point fall detection. While some experts treat the position data as sequences and apply Recurrent Neural Networks (RNN) \cite{b16} and Long Short-Term Memory (LSTM) \cite{b17} for fall detection. Quadros proposed a totally different approach by decomposing the movements by Madgwick’s Decomposition (TBM-MD) and estimating the spatial orientation based on gravity for fall detection \cite{b18}. In this project, we will handle the fall detection problem based on wearable devices which provide sequential position data for further fall detection. The relevant research will be discussed in more details.

Ambient sensors are the general term for vision sensors, sound sensors, radar sensors, infrared sensors and pressure sensors \cite{b19}. In this situation, body movements could be captured by one or several videos via a single (monocular) camera or multiple cameras. Based on these video-based sensors, the fall detection process can be separated into tracking problem to track the human position and classification problem to predict the falls. With monocular depth cameras, body movements could be tracked by Region of Interest (ROI) through background subtraction, and falls could be predicted by setting a appropriate threshold for ROI heights \cite{b20}. In similar scenarios, Bian \cite{b21} proposed an improved randomized decision tree (RDT) to extract 3D body joint trajectory patterns and performed fall detection with a SVM classifier. While in multi-video scenarios, more complex algorithms are involved. In Gomes's research \cite{b22}, YOLO detectors and Kalman Filters are applied to track human movements and falls are detected based on temporal features extracted from 3D-CNN or 2D-CNN + LSTM model. Moreover, Ramirez \cite{b23} introduced a skeleton-based algorithm to detect falls based on human skeleton features extracted by AlphaPose.

As mentioned above, we will mainly focus on fall detection using sequential position data collected by wearable devices in this project. The solutions to sequential-based fall detection can be categorised to three types: supervised binary classification, unsupervised outlier detection and supervised drift detection. 

For supervised binary classification solutions, \cite{b14} and \cite{b15} applied Support Vector Machine (SVM) and Random Forest (RF) to recognize falls according to position feature (numerical values for x, y and z) of each data point. Regarding unsupervised outlier detection, unsupervised methods could be applied, such as Isolation Forest (IS) and Histogram-Based Outlier Score (HBOS) \cite{b24}. Extreme points will be recognized as outliers / falls during the detection. The mentioned two types of point-wise approaches both raise a main limitation that the model would lose contextual information since the point-wise methods completely ignore the temporal dependencies between adjacent points. For instance, position data might be similar whether a person has already fallen or is lying down. 

Aiming to preserve the possible inherent dependencies among neighbouring points, supervised drift detection methods apply sliding window technique to separate the data into multiple segments and perform fall detection based on the trend and drift within the segments. Jain \cite{b25} conducted the hybrid architecture by firstly clustering segments by K-Means to reduce dataset size and applying SVM for drift detection. Torti proposed a 2-layer LSTM architecture to capture the position segments and classify the final results to normal, alert and exact fall \cite{b17}. 

\begin{figure}[htbp]
\centerline{\includegraphics[width=0.5\textwidth]{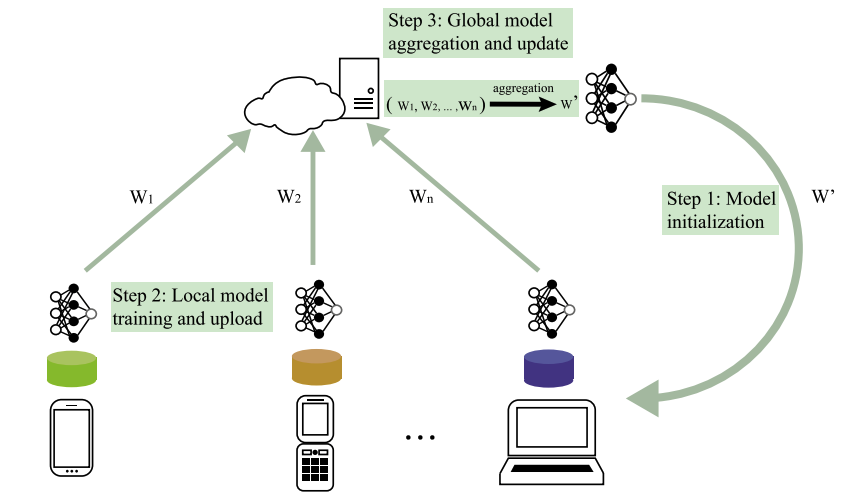}}
\caption{Workflow for Federated Learning\cite{b38}}
\label{fig:1}
\end{figure}

Although segment-based drift detection could improve the model performance by involving inherent dependencies, both point-wise and segment-based approaches above have main concern in user privacy and data security since all the data are integrated together for training without any encryption or pre-processing. The data breaches could lead to significant risks in financial loss or other troubles. Considering this, Federated Learning (FL) could help with the situation by not sharing data among users in a hierarchical way. As figure \ref{fig:1} shows \cite{b26}, several clients will collect their own data, train the model locally and upload their model weights to the server in the middle. The server will aggregated to form a new global model. Then the weights of the global model will be broadcasted to all clients to perform offline inference. In this process, data are decentralized and the security could be ensured without any data transmission.

After obtaining multiple local client models, weight aggregation becomes a crucial topic in FL since vanilla mean aggregation (such as FedAvg\cite{b37}) would lead to very poor performance if the data distribution among different clients varies significantly. Trimmed Mean Aggregation \cite{b34} was invented to remove a fraction of the smallest and largest updates before averaging the remaining ones during the weight aggregation process. The weight updates from client models will be sorted in order and pruned by a trimming factor $\beta$ to remove possible outliers that may cause instability in further training and inference. FedNova\cite{b33} further optimizes the aggregation function by normalizing client updates to account for differences in local training iterations. The more steps a client model trains, the more it shifts. To mitigate this, weight updates are divided by the number of epochs, reducing the impact of these shifts. Apart from optimizing the weight aggregation process, FedProx\cite{b35} adds a penalized loss term to regularize client model training, aiming to prevent client models from drifting too far from the global model.

Regarding the practical usage of FL, \cite{b27} and \cite{b28} introduced hierarchical FL framework with LSTM for fall detection. However, these FL approaches are feasible only when the client devices are capable for local model training. Otherwise, data encryption methodologies should be included before data transmissions between clients and the server, such as TenSEAL \cite{b29} using Homomorphic Encryption (HE) to protect the raw data information. HE provides high-level data encryption and enable data processing without decryption. However, this approach is computational expensive and time consuming comparing to other simpler data encryption methods \cite{b32}.

Even with FL to solve privacy concerns, the all-in-one model could barely generalize behaviours for all users due to the individual differences. Therefore, to address all the limitations mentioned above, we proposed our Ensembled Penalized Federated Learning drift detection model based on sequential position data and will be discussed in detail in the next section.

\section{Data Preprocessing and Baseline Models}
\label{sec:preprocessing}

This subsection outlines the preprocessing steps applied to the raw sensor data collected from wearable devices. As the dataset contains no missing values, we focus on four key preprocessing stages to ensure data quality and model readiness. We begin by detailing the data preprocessing strategies applied to wearable sensor data, followed by an overview of baseline models used for performance benchmarking. 

\subsection{Time-Series Alignment and Segmentation}
The sequential position data are collected simultaneously from multiple wearable sensors, each representing real-time physical movements of an individual. To maintain consistency across sensors and preserve the relative temporal alignment, we standardized the sequence lengths. Specifically, we determined the minimum number of records across all sensors during the same time window and truncated the longer sequences using random sampling while preserving their temporal order.

To enable effective drift detection and temporal modeling, the data were segmented into fixed-length sequences. We considered two common approaches for time-series segmentation: sliding windows and time-fading windows \cite{b30}. The sliding window method retains only the most recent observations, while the time-fading method assigns more weight to recent data while preserving historical context. Given that falls are typically abrupt and not influenced by long-term trends, we selected the sliding window approach, using fixed window size and stride to extract overlapping sub-sequences from the aligned data.

\subsection{Handling Imbalanced Data}
Due to the rarity of fall events, the dataset exhibits significant class imbalance, which can lead to models with high overall accuracy but low sensitivity to falls (i.e., low recall). To address this, we applied the Synthetic Minority Over-sampling Technique (SMOTE) \cite{b31} to synthetically augment the minority class. Unlike random oversampling or majority downsampling, SMOTE generates new, plausible data points in the minority class. To avoid overfitting, we adjusted the class distribution so that fall samples constituted approximately 25\% of the total dataset, rather than fully balancing the classes.

\subsection{Data Standardization} 
Standardization, which transforms features to have zero mean and unit variance, is commonly used to accelerate convergence and improve numerical stability, especially when features vary significantly in scale. However, in our case, all features represent positional coordinates with similar value ranges. Experimental results showed no improvement in performance after standardization; hence, this step was excluded from further processing.

\subsection{Baseline Model Description} \label{sec:baseline}
To benchmark the performance of our proposed federated learning-based framework, we implemented several baseline models that represent traditional approaches to fall detection. These models fall into two main categories: supervised binary classification and unsupervised outlier detection. All baseline models operate in a point-wise fashion, meaning they make predictions based solely on individual time-step records without incorporating temporal dependencies across sequences.

\subsubsection{Supervised Binary Classification Models}  
We selected Support Vector Machine (SVM) and Random Forest (RF) as classical supervised learning algorithms due to their widespread adoption and effectiveness in structured data classification tasks. These models were trained using labeled data, where the input features correspond to positional data from sensors and the output label indicates whether a fall occurred.  

\begin{itemize}
    \item \textit{SVM} attempts to find the optimal hyperplane that separates fall and non-fall instances with the maximum margin. Its robustness to high-dimensional feature spaces and ability to incorporate different kernel functions make it well-suited for early-stage experimentation.
    \item \textit{Random Forest} is an ensemble of decision trees trained on different subsets of the dataset and feature space. Its strength lies in handling non-linear feature interactions and reducing overfitting compared to single decision trees.
\end{itemize}

\subsubsection{Unsupervised Outlier Detection Models}  
For anomaly detection scenarios where explicit labeling may be sparse or noisy, we employed Isolation Forest (IF) and Histogram-Based Outlier Score (HBOS) as representatives of unsupervised learning.

\begin{itemize}
    \item \textit{Isolation Forest} identifies anomalies by isolating observations through randomly generated trees. Anomalous points require fewer splits to isolate, making them distinguishable from normal instances.
    \item \textit{HBOS} computes histograms for each feature and assigns an outlier score based on deviations from the expected distribution. It is highly scalable and well-suited for real-time anomaly detection.
\end{itemize}

\subsection{Model Selection and Tuning}  
Each baseline model was fine-tuned using grid search over a predefined hyperparameter space to maximize the recall score—a critical metric for fall detection, where missing a true fall could have serious consequences. Cross-validation was performed during the tuning process to ensure robustness and generalizability of the models. Table~\ref{tab:grid} summarizes the grid search configurations used for each baseline model.

\begin{table}[htbp]
\caption{Grid Search Settings for Baseline Models}
\begin{center}
\begin{tabular}{|c|l|}
\hline
Model & Hyperparameters and Search Range \\ 
\hline
SVM & \makecell[l]{C: \{0.001, 0.01, 0.1, 1.0\} \\
      Kernel: \{'linear', 'poly', 'rbf'\} \\
      Class weight: \{None, 'balanced'\}} \\
\hline
RF & \makecell[l]{n\_estimators: \{100, 150, 200, 250, 300\} \\
      max\_depth: \{5--15\} \\
      min\_samples\_split: \{2, 4, 6, 8, 10\} \\
      min\_samples\_leaf: \{1--10\} \\
      Class weight: \{None, 'balanced', 'balanced\_subsample'\}} \\
\hline
IF & \makecell[l]{n\_estimators: \{100, 150, 200, 250, 300\} \\
      max\_features: \{0.5, 0.6, 0.7, 0.8, 0.9, 1.0\}} \\
\hline
HBOS & \makecell[l]{Threshold: \{0.5, 0.6, 0.7, 0.8, 0.9\}} \\
\hline
\end{tabular}
\label{tab:grid}
\end{center}
\end{table}

After conducting grid search, we selected the optimal hyperparameters for each model based on the best-performing configuration for two dataset types: Single Subject Dataset (SSD) and Multi-Subject Dataset (MSD) (described in Section IV). These fine-tuned parameters, which were subsequently used in our final baseline evaluations, are summarized in Table~\ref{tab:baseline_settings}.

\begin{table}[htbp]
\caption{Fine-Tuned Hyperparameters for Baseline Models}
\renewcommand{\arraystretch}{1.2}
\begin{center}
\begin{tabular}{|c|c|c|l|}
\hline
ID & Model & Dataset & Hyperparameters \\
\hline
1 & SVM & SSD & \makecell[l]{C=1.0, kernel='rbf', \\ class\_weight='balanced'} \\
\hline
2 & SVM & MSD & \makecell[l]{C=1.0, kernel='rbf', \\ class\_weight='balanced'} \\
\hline
3 & RF & SSD & \makecell[l]{n\_estimators=200, max\_depth=10, \\ min\_samples\_split=5, min\_samples\_leaf=2, \\ class\_weight='balanced'} \\
\hline
4 & RF & MSD & \makecell[l]{n\_estimators=200, max\_depth=10, \\ min\_samples\_split=5, min\_samples\_leaf=10, \\ class\_weight='balanced'} \\
\hline
5 & IF & SSD & \makecell[l]{n\_estimators=300, max\_sample='auto', \\ contamination='auto', max\_features=1.0} \\
\hline
6 & IF & MSD & \makecell[l]{n\_estimators=300, max\_sample='auto', \\ contamination='auto', max\_features=0.5} \\
\hline
7 & HBOS & SSD & \makecell[l]{threshold=0.5} \\
\hline
8 & HBOS & MSD & \makecell[l]{threshold=0.5} \\
\hline
\end{tabular}
\label{tab:baseline_settings}
\end{center}
\end{table}

While these baseline models serve as strong references, their inherent limitations—namely the lack of temporal modeling and centralized training assumptions—underscore the need for our proposed federated and sequence-aware architecture tailored to real-world, privacy-sensitive, and time-dependent biomedical applications.

\section{Proposed Ensembled Penalized Federated Learning (EPFL) Framework}

\begin{figure*}[t]
    \centering
    \begin{subfigure}[b]{0.75\textwidth}
        \includegraphics[width=\textwidth]{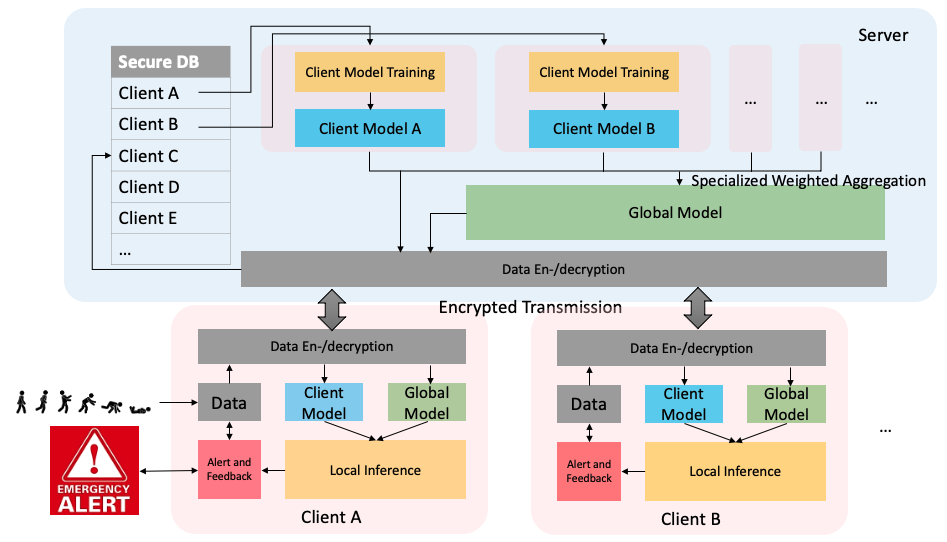}
        \caption{Overall workflow of the proposed Ensembled Penalized Federated Learning framework, illustrating the process of client-side training, aggregation using Specialized Weighted Aggregation (SWA), and ensemble integration.}

        \label{fig:subfig1}
    \end{subfigure}
    \hfill
    \begin{subfigure}[b]{0.18\textwidth}
        \includegraphics[width=\textwidth]{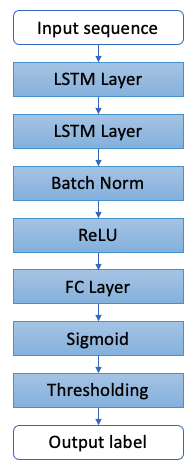}
        \caption{Detailed architecture of the LSTM-based client model used for sequential fall detection.}
        \label{fig:subfig2}
    \end{subfigure}
    \caption{Illustration of the proposed model architecture. (a) Describes the complete federated learning workflow with ensemble and penalization components. (b) Shows the internal architecture of the client-side sequential detection model.}
    \label{fig:mainfig}
\end{figure*}

This section presents the proposed \textit{Ensembled Penalized Federated Learning} (EPFL) framework for privacy-preserving and personalized fall detection. EPFL integrates federated learning, robust aggregation, ensemble inference, and user feedback to improve sensitivity while preserving user data confidentiality. The system comprises a centralized server and multiple distributed clients equipped with wearable sensors, as illustrated in Figure~\ref{fig:mainfig}.

Let $\mathcal{D}_i = \{(x^{(i)}_t, y^{(i)}_t)\}_{t=1}^{T}$ denote the private dataset for client $i$, where $x^{(i)}_t$ represents sequential motion sensor data and $y^{(i)}_t \in \{0, 1\}$ indicates a fall event. The global training objective is:
\begin{equation}
\min_{w} \sum_{i=1}^N \frac{1}{|\mathcal{D}_i|} \sum_{t=1}^{T} \ell(f(x^{(i)}_t; w), y^{(i)}_t) + \mu \| w - w^g \|^2
\end{equation}
subject to the constraint that raw data and model updates are transmitted only in encrypted form.

In this architecture, each client collects local sequential motion data using wearable devices. Before transmission, the data are encrypted to prevent privacy breaches. Once received, the server decrypts and stores the data in a secure environment and uses it to train client-specific models. These models are aggregated through a proposed \textit{Specialized Weighted Aggregation} (SWA) technique to form an updated global model. This updated model, along with each client model, is re-encrypted and distributed back to the respective clients. Inference is then performed using an ensemble of global and client-specific predictions. Finally, a feedback mechanism allows for continual model improvement based on user input regarding fall alerts.

\subsection{Data Encryption}
To address privacy concerns, all motion data and model parameters transmitted between clients and the server are protected using \textit{Homomorphic Encryption} (HE). Let $E(\cdot)$ denote the encryption function. Each client transmits $E(w_i)$ to the server, enabling aggregation without decryption:
\begin{equation}
E(w_g) = \text{Aggregate}(E(w_1), E(w_2), ..., E(w_N))
\end{equation}. This cryptographic method enables computations on encrypted data without requiring decryption, thereby preserving confidentiality throughout the communication process.

In this implementation, encryption is applied only during data and model transmission, not during local model training or inference, due to the significant computational overhead of HE. Transmissions are scheduled periodically rather than in real-time to minimize communication and processing delays. The HE protocol is implemented using the TenSEAL Python library, which supports secure tensor computations.

\subsection{Client and Global Model Architecture}
Each model in the system—both global and client-specific—is designed using a deep learning architecture based on \textit{Long Short-Term Memory} (LSTM) networks to capture the temporal dynamics of sensor data. As illustrated in Figure~\ref{fig:subfig2}, the architecture consists of two stacked LSTM layers, followed by batch normalization, ReLU activation, a fully connected layer, and a sigmoid output layer for binary classification. 

Each client and the server deploy an identical LSTM-based architecture. The model is formally defined as:
\begin{equation}
h_t = \text{LSTM}(x_t, h_{t-1}); \quad z = \sigma(W_2 \cdot \text{ReLU}(W_1 h_T + b_1) + b_2)
\end{equation}
where $x_t$ is the sensor input, $h_t$ is the LSTM hidden state, and $z$ is the predicted fall probability. The final binary decision is made using:
\begin{equation}
\hat{y} = \mathbb{I}(z > threshold)
\end{equation}

The model outputs a fall probability score, which is thresholded (grid search between 0.3 and 0.5 to find the best performance, set to 0.3 in EPFL) to determine the final binary prediction. This threshold was empirically selected to maximize recall while minimizing false positives in the context of class imbalance.

\subsection{Proposed Specialized Weighted Aggregation (SWA)}
One of the core contributions of our proposed framework is the \textit{Specialized Weighted Aggregation} (SWA) strategy, which enhances federated model robustness in the presence of heterogeneous client behavior and noisy updates. Unlike traditional aggregation methods like FedAvg, SWA integrates normalization, outlier-resistant averaging, and weighted global update fusion to ensure stable and fair learning dynamics.

The proposed Specialized Weighted Aggregation (SWA) mechanism addresses several limitations inherent in conventional FedAvg. First, it adapts to varying local training efforts by normalizing each client's update based on the number of training epochs, ensuring fairer contributions from heterogeneous clients. Second, by employing trimmed mean aggregation, SWA effectively filters out extreme or noisy updates, which are common in real-world scenarios involving wearable devices with inconsistent data quality or sensor drift. Lastly, the use of exponential moving average for global update fusion enhances stability and convergence, reducing abrupt shifts in model performance across communication rounds. These advantages make SWA particularly suitable for healthcare applications, where device reliability, user behavior, and environmental conditions can vary significantly between individuals.

\subsubsection{FedNova Normalization}
To mitigate the effect of unequal training efforts across clients, we normalize each client update by the number of local training epochs. Let $w_i$ denote the model update from client $i$ and $e_i$ the number of local epochs. The normalized update is computed as:
\begin{equation}
    w'_i = \frac{w_i}{e_i}
\end{equation}
This step prevents disproportionately large updates from clients with longer training durations.

\subsubsection{Trimmed Mean Aggregation}
To increase resilience against noisy, faulty, or adversarial updates, we apply a trimmed mean strategy. First, we sort the set of normalized updates in ascending order for each dimension. Then, we discard the top and bottom $m$ updates, where:
\begin{equation}
    m = 
    \begin{cases}
    \left\lfloor \beta \cdot n \right\rfloor &,\  \text{if } \left\lfloor \beta \cdot n \right\rfloor \geq 1 \\
    1 &,\ \text{otherwise}
    \end{cases}
\end{equation}
Here, $\beta$ is a trimming factor (set to 0.1 in our experiments), and $n$ is the total number of clients. Following \cite{b34}, the aggregated update of n 1-dimensional scalers is stated below after we sort the scalers in ascending order:
\begin{equation}
    Tmean(u_i, u_2, ..., u_n) = \overline{u}_{\text{trim}} = \frac{1}{n - 2m} \sum_{i = m+1}^{n - m} u_{i}
\end{equation}

In our project, each $w'_i$ is a d-dimensional vector and we need to calculate the aggregated update for each dimension from the remaining $n - 2m$ clients which also follows the formula in \cite{b34}:
\begin{equation}
    \overline{w}_{\text{trim}} = \sum_{i=1}^{d} e_i\cdot Tmean(<w'_1, e_i>, ..., <w'_n, e_i>)
\end{equation}
where $e_i$ is the i-th column of the $d\times d$ identity matrix. 

\subsubsection{Global Weighted Fusion}
Instead of directly replacing the global model with the new average, we apply a soft update using exponential moving average fusion. This smooths the update process and reduces oscillations in global parameters:
\begin{equation}
    w_g^{t+1} = (1 - \alpha) w_g^t + \alpha \cdot \overline{w}_{\text{trim}}^{t+1}
\end{equation}
where $\alpha$ is the fusion weight, empirically set to 0.1, and $w_g^t$ represents the global model at round $t$.

\subsubsection{Complete SWA Procedure}

The complete aggregation process in our Specialized Weighted Aggregation (SWA) strategy. This procedure ensures robust global model updates by mitigating the effects of variable local training effort, eliminating extreme updates, and applying a stabilized fusion scheme.

First, each client's model update is normalized by its number of local training epochs to account for computational heterogeneity. The normalized updates are then sorted based on their distance from the mean update. A trimming operation removes the top and bottom $\beta$ fraction of updates, which helps suppress noisy or potentially adversarial contributions. Finally, the trimmed mean of the remaining updates is computed and softly integrated into the global model using an exponential moving average with a tunable fusion weight $\alpha$.

\subsection{Personalized Local Ensemble Inference}

To account for inter-individual variability in motion patterns and enhance prediction personalization, we adopt a local ensemble inference strategy. Each client maintains both the global model $f_g(x)$, which generalizes across all clients, and a locally trained model $f_i(x)$, which captures user-specific features. During inference, the final fall probability $\hat{y}$ is computed by averaging the predictions from both models:

\begin{equation}
    \hat{y} = \frac{1}{2} \left( f_g(x) + f_i(x) \right)
\end{equation}

This dual-model inference approach enables the system to leverage both broad, population-level knowledge and fine-tuned personal behavior patterns. Such a hybrid strategy is particularly beneficial in healthcare scenarios where physiological signals and movement signatures can vary significantly across individuals.

\subsection{Alert Generation and Feedback-Driven Adaptation}

An emergency alert is triggered if the ensemble fall probability $\hat{y}$ exceeds a predefined threshold $\theta$ (set empirically at 0.4). The alert mechanism is designed to minimize false negatives, ensuring timely intervention in high-risk situations.

To further improve reliability and adaptiveness, we incorporate a user-driven feedback loop. After each alert, the user is prompted to confirm or reject the event. This feedback $r_t \in \{0, 1\}$ is used to correct or reinforce the corresponding label, and the refined instance $(x_t, r_t)$ is added to the client’s local dataset:

\begin{equation}
    \mathcal{D}_i^{(t+1)} = \mathcal{D}_i^{(t)} \cup \{(x_t, r_t)\}
\end{equation}

These periodically updated labels are used in future training rounds, allowing the model to continuously adapt to evolving behavior patterns and sensor characteristics. This human-in-the-loop mechanism not only improves prediction accuracy over time but also builds user trust in the system's decisions.


\subsection{Model Training and Evaluation Setup}
To ensure consistency across clients, all models share identical hyperparameters: hidden size of 128, batch size of 32, learning rate of 0.001, and the Adam optimizer. The global model is trained for 60 epochs, while client models train for 30 epochs. Early stopping is employed by monitoring the sum of Recall and F1-score on the validation set to prevent overfitting.

The training loss function integrates \textit{Binary Cross-Entropy} (BCE) with a FedProx regularization term to discourage local models from drifting too far from the global representation. The combined objective function is expressed as:

\begin{equation}
L = L_{\text{BCE}} + L_{\text{FedProx}}
\end{equation}
\begin{equation}
L = -\left[y \cdot \log(\hat{y}) + (1 - y) \cdot \log(1 - \hat{y})\right] + \mu \sum_{i=1}^{d} \| w_i - w^g_i \|^2
\end{equation}

where $y$ is the ground truth label, $\hat{y}$ the predicted probability, $w_i$ the local model weights, $w^g_i$ the corresponding global weights, and $\mu$ is a tunable penalty factor (set to 0.01).

For evaluation, the dataset is partitioned into training and test sets. Only the test set is used for performance assessment. We compute Accuracy, Precision, Recall, and F1-score, prioritizing Recall due to the critical nature of detecting fall events and minimizing false negatives.

\section{Experiment and Result Analysis}
This section presents an evaluation of the proposed EPFL framework for fall detection in smart home environments. We describe the dataset, preprocessing steps, evaluation metrics, and experimental setup. We then compare the performance of traditional classifiers, outlier detectors, and sequential models under centralized and federated learning settings. Finally, we analyze convergence behavior, the impact of preprocessing and oversampling, and the role of ensemble and personalized aggregation in enhancing robustness and generalization.

\subsection{Dataset Description}
The dataset used in this study originates from LDPA\cite{b36}, originally collected for developing safer smart environments, specifically for fall detection among elderly individuals within an independent smart home setting. It consists of 7 features: sequence name, sensor tag, timestamp, date, 3D positional data (X, Y, Z). Data from 5 individuals (A, B, C, D and E) is recorded, with each individuals performing 5 activity sequences. The sequence name (E02 for the second activity sequence of individual E) is used to identify each sequence. The sensor tag indicates the body part where the sensor is placed (table~\ref{tab:1}).
\begin{table}[htbp]
\caption{Sensors for different body parts}
\begin{center}
\begin{tabular}{|c|c|}
\hline
Sensor ID & Body Location \\
\hline
010-000-024-033 & Left Ankle \\
010-000-030-096 & Right Ankle \\
020-000-033-111 & Chest \\
020-000-032-221 & Belt \\
\hline
\end{tabular}
\label{tab:1}
\end{center}
\end{table}

Each data record is labeled with an activity status and the distribution of activities is shown in table~\ref{tab:2}. The table indicates that falling events (anomalies) are significantly underrepresented, leading to a highly imbalanced class distribution between normal and anomalous events. 

\begin{table}[htbp]
\caption{Activity status in LDPA}
\begin{center}
\begin{tabular}{|l|c|c|}
\hline
Status & Amount & Proportion \\
\hline
lying & 54480 & 33.05\% \\
\hline
walking & 32710 & 19.84\% \\
\hline
sitting & 27244 & 16.53\% \\
\hline
standing up from lying & 18361 & 11.14\% \\
\hline
sitting on the ground & 11779 & 7.14\% \\
\hline
lying down & 6168 & 3.74\% \\
\hline
on all fours & 5210 & 3.16\% \\
\hline
falling & 2973 & 1.80\% \\
\hline
standing up from sitting on the ground & 2848 & 1.73\% \\
\hline
sitting down & 1706 & 1.03\% \\
\hline
standing up from sitting & 1381 & 0.84\% \\
\hline
\end{tabular}
\label{tab:2}
\end{center}
\end{table}

A refactored version of LDPA is available on Kaggle (Anomaly Detection Falling People Events) with timestamps already removed and activities relabeled to 0 (normal) and 1 (falling). This version is used for point-wise model experiments as mentioned in \ref{sec:baseline}. For both point-wise models and sequential models, we allocate 4 activity sequences per individual for training and 1 sequence for testing. Two preprocessing methods were applied in this study.

\subsubsection{Separated Sensor Data (SSD)}
In this approach we directly utilized the raw dataset based on the splitted training and test data. Only the 3-dimensional position data and the corresponding activity label will be used for training and inference. Due to the lack of sequence alignment, only point-wise models will apply this approach.

\subsubsection{Merged Sensor Data (MSD)}
Referring to section ~\ref{sec:preprocessing}, we applied Time-Series Alignment to merge the position data from different sensors together. Considering the high similarity between two ankles, we only take one of them with more data records. Then we align the records from three sensors to form a new format of 9 features, including ‘ANKLE\_x', ‘ANKLE\_y’, ‘ANKLE\_z’, 'CHEST\_x', 'CHEST\_y', 'CHEST\_z', 'BELT\_x', 'BELT\_y', 'BELT\_z'. Finally, we relabel the data records to 1 if any of ankle, chest and belt holding the activity label "falling", and 0 otherwise. Moreover, we additionally use the sliding window (window\_size=20, stride=1) to segment the data series for data used in sequential models.

\subsection{Evaluation Metrics}
Several commonly used evaluation metrics for classification are used to evaluate the model performance, including Accuracy, Precision, Recall, and F1-score.

\begin{equation}
    \text{Accuracy} = \frac{TP + TN}{TP + TN + FP + FN}
\end{equation}

\begin{equation}
    \text{Precision} = \frac{TP}{TP + FP}
\end{equation}

\begin{equation}
    \text{Recall} = \frac{TP}{TP + FN}
\end{equation}

\begin{equation}
    \text{F1-score} = \frac{2 \times \text{Precision} \times \text{Recall}}{\text{Precision} + \text{Recall}}
\end{equation}

Given the background of this dataset, among the commonly used evaluation metrics, Recall is the most crucial. Since our primary goal is anomaly detection, specifically identifying falling incidents among elderly individuals, we prioritize detecting as many anomalies as possible. Therefore, we aim to maximize Recall, even at the cost of lower Accuracy and Precision.

\subsection{Software and Hardware Requirements}
The project is running on CPU-only machine with 16G Memory. Several third-party packages are installed based on Python 3.8 in Anaconda, including numpy, pandas, scikit-learn, torch, matplotlib, seaborn and tenseal. Please note that tenseal can only support Python version $\geq$ 3.8.

\begin{table*}[htbp]
\caption{Comparison of Model Performance Across Classification Tasks: Binary Classification (BC), Outlier Detection (OD), and Distributed Detection (DD). Metrics include Accuracy, Precision, Recall, F1-score, and Class-wise Recall (A–E) where applicable.}

\begin{center}
\begin{tabular}{|c|c|c|c|c|c|c|c||c|c|c|c|c|}
\hline
Task & ID &  Model & Settings & Acc. & Prec. & Rec. & F1-score & Rec.-A & Rec.-B & Rec.-C & Rec.-D & Rec.-E \\
\hline
BC & 1 & SVM & SSD & 0.7210 & 0.1414& 0.8286& 0.2416&- &- &- &- &- \\
BC & 2 & SVM & MSD & 0.7793& 0.2364& 0.4931& 0.3196&- &- &- &- &- \\
BC & 3 & RF & SSD & 0.7640& 0.1259& 0.5727& 0.2065&- &- &- &- &-\\
BC & 4 & RF & MSD & 0.8277& 0.2780& 0.4003& 0.3282&- &- &- &- &-\\
OD & 5 & IF & SSD & 0.4442& 0.0595& 0.6329& 0.1088&- &- &- &- &- \\
OD & 6 & IF & MSD & 0.5647& 0.1097& 0.4416& 0.1758&- &- &- &- &- \\
OD & 7 & HBOS & SSD & 0.0795& 0.0529& 0.9565& 0.1002&- &- &- &- &- \\
OD & 8 & HBOS & MSD & 0.1051& 0.1051& \textbf{1.0000}& 0.1902&- &- &- &- &- \\
\hline
DD & 9 & LSTM & MSD & 0.9852& 0.8107& 0.8741& 0.8412&- &- &- &- &- \\
DD & 10 & LSTM & SMOTE+MSD & 0.9852& 0.7806& 0.9022& 0.8370&- &- &- &- &- \\
DD & 11 & LSTM & FL+MSD & 0.2818& 0.0439& 0.9153& 0.0838& 0.4375&0.8485 &0.5333 & 0.5690& 0.5435\\
DD & 12 & LSTM & PFL+SWA+MSD & \textbf{0.9928}& \textbf{0.9195}& 0.8750& 0.8967& 0.5312& 0.7879& 0.3111& 0.4655& 0.6522\\
DD & 13 & LSTM & EPFL+SWA+MSD & \textbf{0.9929}& \textbf{0.9163}& 0.8831& \textbf{0.8994}& \textbf{0.8444}& \textbf{1.0000}& \textbf{0.8222}& \textbf{0.9114}& \textbf{0.8478}\\
\hline

\end{tabular}
\label{tab:result}
\end{center}
\end{table*}

\subsection{Results and Discussion}

This section offers a detailed evaluation of the proposed Ensembled Penalized Federated Learning (EPFL) framework for fall detection, benchmarked against both classical and contemporary baselines. The evaluation spans three task categories: Binary Classification (BC), Outlier Detection (OD), and Distributed Detection (DD), as reported in Table~\ref{tab:result}. The metrics under consideration include Accuracy, Precision, Recall, and F1-score, with an emphasis on Recall due to its criticality in fall detection applications, where failure to detect a fall may lead to catastrophic consequences.

\subsubsection{Performance of Individual Models}

The performance of each model was evaluated using standard classification metrics—\textit{Accuracy}, \textit{Precision}, \textit{Recall}, and \textit{F1-score}—to comprehensively assess their effectiveness in fall detection. The classical machine learning models, \textit{Support Vector Machine (SVM)} and \textit{Random Forest (RF)}, demonstrated moderately high accuracy but lower precision and F1-score, a common issue when dealing with imbalanced data.

Specifically, under the Multi-Sensor Data (MSD) setting, \textit{SVM} achieved an Accuracy of 77.93\%, a Precision of 23.64\%, a Recall of 49.31\%, and an F1-score of 31.96\%. Similarly, \textit{RF} yielded an Accuracy of 82.77\%, Precision of 27.80\%, Recall of 40.03\%, and F1-score of 32.82\%. These numbers confirm that while both SVM and RF can classify normal events well, they struggle with fall detection (minority class), leading to reduced F1-scores and limiting their practical application.

In contrast, the sequential model, \textit{Long Short-Term Memory (LSTM)}, significantly outperformed traditional classifiers across all metrics. The LSTM trained on MSD achieved an Accuracy of 98.52\%, Precision of 81.07\%, Recall of 87.41\%, and an F1-score of 84.12\%. These results indicate a strong balance between correctly identifying actual falls and avoiding false alarms. The high Recall reflects LSTM's effectiveness in capturing fall-related patterns, while the high Precision shows it does not compromise significantly on false positives.

The improved performance of LSTM can be attributed to its ability to model the temporal dependencies in the sensor data. Unlike traditional classifiers, which rely on static feature representations, the LSTM network learns sequential patterns from wearable sensor data, enabling it to distinguish dynamic transitions associated with fall events. These results confirm that \textit{incorporating temporal sequence modeling substantially improves fall detection sensitivity}.

In addition to supervised models, we also examined unsupervised outlier detection approaches: \textit{Isolation Forest} and \textit{Histogram-Based Outlier Score (HBOS)}, treating falls as anomalous behavior in daily activity patterns. For instance, HBOS under MSD achieved a perfect Recall of 100.00\%, but a very low Precision of 10.51\%, resulting in an F1-score of only 19.02\%. Isolation Forest, under MSD, showed a Recall of 44.16\%, Precision of 10.97\%, and F1-score of 17.58\%. These values reflect the high false-positive rates of unsupervised methods.

These results reveal a clear trade-off: \textit{unsupervised outlier detection models favor high sensitivity (recall) at the cost of precision}, resulting in frequent false alarms. This limitation reduces their viability for real-world deployment, where too many false alerts could reduce user trust and system utility. The significantly lower F1-scores of Isolation Forest and HBOS, compared to the LSTM model, support this conclusion.

Therefore, the results \textit{justify the use of supervised and temporal models}, such as LSTM, as the core approach for fall detection, particularly when annotated datasets are available. The LSTM’s superior performance across all four evaluation metrics underlines its suitability for real-time, reliable fall detection applications, balancing sensitivity with false alarm minimization.

\begin{figure}[htbp]
\centerline{\includegraphics[width=0.45\textwidth]{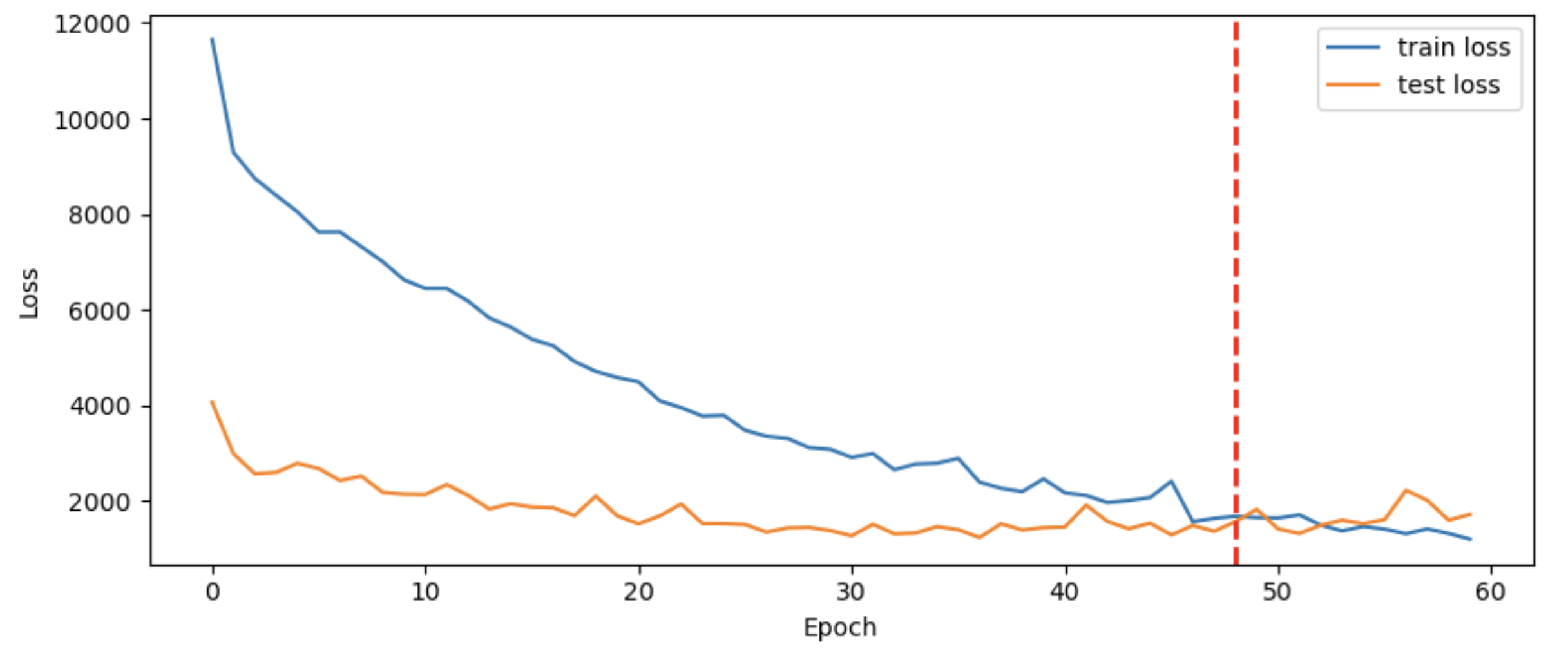}}
\caption{Training and validation loss over epochs for the sequential LSTM model. The plot demonstrates stable convergence, with the test loss flattening around epoch 45, indicating optimal early stopping.}
\label{fig:loss}
\end{figure}

\subsubsection{Convergence Behavior and Training Stability of Baseline Model}

We further analyzed the convergence behavior and training stability of the proposed model. Fig.~\ref{fig:loss} illustrates the training and test loss trajectories of the sequential LSTM model over 60 epochs.

As shown in the figure, the training loss exhibits a consistent downward trend, gradually decreasing from above 11,000 to below 2,000. The test loss also follows a similar pattern, initially dropping rapidly and then stabilizing around epoch 45. After this point, test loss fluctuations increase slightly, suggesting potential overfitting if training continues further. A red dashed line at epoch 47 marks the early stopping point, where the test loss is lowest and most stable.

This convergence behavior indicates that the model effectively learns the underlying patterns in the data within the first 45 epochs, and training beyond this point does not yield significant performance gains. The early stopping mechanism based on validation loss thus prevents overfitting and ensures better generalization on unseen data.

Overall, the figure demonstrates stable convergence and effective training dynamics for the centralized LSTM model, validating its suitability as a baseline for further comparison with federated variants.

\subsubsection{Impact of Preprocessing and Oversampling Techniques}

We evaluated each model's performance under two preprocessing setups—\textit{Single-Sensor Data (SSD)} and \textit{Multi-Sensor Data (MSD)}—to understand the impact of sensor fusion. The MSD configuration combined accelerometer and gyroscope data streams, whereas SSD relied solely on a single sensor modality. Across all models, the MSD setup consistently led to superior performance.

For instance, the SVM model improved from an F1-score of 0.2416 under SSD (ID 1) to 0.3196 under MSD (ID 2), with Accuracy increasing from 72.10\% to 77.93\%. Similarly, RF’s F1-score rose from 0.2065 (ID 3) to 0.3282 (ID 4), and Accuracy from 76.40\% to 82.77\%. While these gains are moderate for traditional classifiers, the improvement is more pronounced in the sequential LSTM model. Specifically, LSTM achieved an F1-score of 0.8412 under MSD (ID 9), significantly outperforming the point-wise models and indicating its ability to extract richer temporal features from multimodal input.

These results are expected since gyroscope data (angular velocity) complements accelerometer data (linear acceleration), enabling more robust modeling of the nuanced motion dynamics characteristic of fall events. The added sensor stream helps detect subtle postural transitions or rotational movements that might be invisible to a single modality. In practical applications, this translates to increased reliability—especially in edge cases where falls may manifest weakly across only one sensor type. Despite the increased computational complexity associated with processing multiple streams, the substantial performance boost across models—especially for LSTM—justifies the use of multi-sensor fusion. Consequently, all federated learning models in the later stages of our study were developed using MSD for maximum effectiveness.

To mitigate the significant class imbalance between fall and non-fall events, we employed SMOTE during training to synthetically augment minority-class samples. This oversampling technique was particularly evaluated on the LSTM model under MSD.

As observed in Table~\ref{tab:result}, LSTM without SMOTE (Model 9) achieved a Recall of 0.8741, Precision of 0.8107, and F1-score of 0.8412. When SMOTE was applied (Model 10), Recall improved to 0.9022, reflecting a higher sensitivity to fall events. However, this enhancement came at the cost of Precision, which dropped to 0.7806, and a slight reduction in F1-score to 0.8370. This trade-off highlights the typical behavior of oversampling: while the model becomes more adept at detecting true positives (falls), it also becomes more susceptible to false positives due to synthetic examples that may lie near the decision boundary.

Despite the marginal drop in F1-score, the increased Recall can be beneficial in scenarios where false negatives (missed falls) are far more critical than occasional false alarms. Nevertheless, considering the need to balance high sensitivity with practicality (i.e., avoiding alarm fatigue), we selected the non-SMOTE model (Model 9) as the base for federated learning optimization. This decision reflects a design philosophy that prioritizes generalization and stability, while still ensuring a strong recall rate.

In summary, the use of MSD improves the model’s input representation space and overall detection robustness, while SMOTE selectively boosts sensitivity at the cost of precision. Together, these techniques provide a valuable toolkit for addressing the dual challenges of data sparsity and imbalance in real-world fall detection systems.

\subsubsection{Centralized vs Federated vs Personalized Learning}

We evaluated the LSTM-based model under three different learning paradigms: centralized training, standard federated learning (FL), and our proposed personalized federated strategy. The goal was to investigate how privacy-preserving decentralized techniques compare to the performance upper bound offered by centralized learning.

\textbf{Centralized Training:} In the centralized setup, where data from all clients is aggregated on a single server, the LSTM model (Model 9) achieved the best overall performance, serving as an upper-bound reference. As shown in Table~\ref{tab:result}, this model achieved an Accuracy of 98.52\%, Precision of 0.8107, Recall of 0.8741, and an F1-score of 0.8412. The ability to train on a fully unified dataset allowed the model to capture diverse movement patterns and rare fall signatures comprehensively. However, despite its performance advantages, centralized training is impractical in real-world applications due to stringent privacy constraints, data ownership concerns, and limited communication bandwidth—especially in healthcare systems where data is sensitive and distributed across personal devices.

\textbf{Standard Federated Learning:} To address these constraints, we implemented a standard FL framework using FedAvg aggregation with the same LSTM architecture (Model 11). This model achieved a Recall of 0.9153, which interestingly surpassed that of the centralized version, indicating high sensitivity to fall events. However, this improvement came at the cost of Precision (0.0439) and a dramatically reduced F1-score (0.0838). This drop illustrates the challenge of model drift and performance inconsistency due to data heterogeneity across clients. Individual client recall scores varied significantly—ranging from 0.4375 to 0.8485—highlighting the limitations of using a uniform global model for all users.

\textbf{Personalized Federated Learning (EPFL+SWA):} To mitigate this issue, we proposed an ensemble-based personalized federated framework—\textit{Ensembled Penalized Federated Learning with Specialized Weighted Aggregation (EPFL+SWA)}. This strategy aims to balance global knowledge sharing with local specialization. The EPFL+SWA model (Model 13) achieved the highest performance across distributed approaches, with an Accuracy of 99.29\%, Precision of 0.9163, Recall of 0.8831, and F1-score of 0.8994. More importantly, per-client recall performance was markedly improved, ranging from 0.8222 to 1.0000, demonstrating consistent detection capability across diverse user profiles.

The key innovations in EPFL+SWA—penalizing divergence in local model updates and introducing client-specific weighted aggregation—allow the model to adapt more effectively to the unique data distribution of each client. Unlike meta-learning-based personalization, which requires fine-tuning after training, our method integrates personalization into the training loop, reducing overhead while enhancing adaptability. The ensemble mechanism further stabilizes the aggregated model by capturing multiple local perspectives.

\textbf{Comparative Summary:} Compared to standard FL (Model 11), the EPFL+SWA model (Model 13) improved Precision by more than 87 percentage points (from 0.0439 to 0.9163) and F1-score by over 81 points (from 0.0838 to 0.8994), while maintaining high Recall. These improvements reflect not only better fall detection capability but also reduced false alarm rates—critical for real-world deployment where alarm fatigue is a major concern.

In summary, EPFL+SWA achieved a favorable trade-off between sensitivity and specificity, delivering performance that closely approached or even exceeded the centralized benchmark, while maintaining full data decentralization. This confirms the efficacy of our personalized federated learning strategy as a privacy-preserving, high-accuracy alternative to traditional centralized learning for fall detection in distributed environments.

\subsubsection{Ensemble and Specialized Weighted Aggregation}
The selection of the EPFL+SWA model was the result of a rigorous evaluation of trade-offs among classification metrics in the context of fall detection—a safety-critical, imbalanced classification task. In such applications, \textit{Recall} (sensitivity) is of paramount importance since missing a fall can have severe consequences, while a false alarm, though undesirable, is less critical. Therefore, our model selection prioritized Recall while maintaining acceptable levels of Precision to reduce false positives.

The ensemble nature of EPFL+SWA significantly contributed to the model’s superior performance. By aggregating predictions from multiple local models trained across heterogeneous client datasets, the ensemble approach reduced variance and improved robustness. Specifically, the EPFL+SWA model (Model 13) attained a global Recall of 0.8831, Precision of 0.9163, and an F1-score of 0.8994, outperforming all other federated configurations, including standard FedAvg (Model 11), which suffered from a high Recall (0.9153) but extremely low Precision (0.0439), leading to a degraded F1-score (0.0838).

The ensemble aspect mitigates the “single point of failure” issue common in monolithic models. For instance, if one local model misclassifies a fall, others in the ensemble can compensate, thereby improving overall Recall. The specialized weighting in aggregation further enhances performance by prioritizing client updates that demonstrated better local detection performance or had more representative data. This weighting strategy helped prevent unreliable or noisy updates from skewing the global model, thereby preserving both Precision and generalizability.

The penalization term in EPFL serves a regularization function akin to FedProx. It discourages excessive divergence from the global model during local updates, thus constraining each client’s optimization path within a well-behaved region of the loss landscape. This stabilization not only improves model convergence but also reduces the likelihood of overfitting to local data idiosyncrasies—particularly important when client data distributions are highly non-IID.

The impact of these design choices is quantitatively validated in our results: EPFL+SWA (Model 13) improved per-client Recall scores substantially compared to Model 11 (FedAvg). For example, Recall-A increased from 0.4375 to 0.8444, and Recall-D rose from 0.5690 to 0.9114. Such consistency across clients highlights the effectiveness of the ensemble-weighted strategy in providing both global generalization and personalized specialization. In comparison, simpler models such as SVM or RF (Models 1–4) achieved Recall in the range of 0.40–0.82 but with significantly lower F1-scores, underscoring their inadequacy for this high-stakes application.

Thus, EPFL+SWA offers a principled solution that balances the competing demands of high Recall, acceptable Precision, and robustness across heterogeneous clients, making it the most suitable architecture for privacy-preserving, distributed fall detection systems.

\section{Conclusion and Future Work}
In this study, we proposed an \textit{Ensembled Penalized Federated Learning} (EPFL) framework for fall detection, incorporating a novel \textit{Specialized Weighted Aggregation} (SWA) strategy to effectively combine client models. Experimental results demonstrated that all sequential models significantly outperformed traditional point-wise baseline models. Our proposed EPFL+SWA model achieved superior performance, with a Recall of 0.8831 and an F1-score of 0.8994, indicating its strong ability to detect falls accurately while maintaining a low false alarm rate. Moreover, the decentralized nature of the model supports data privacy by avoiding centralized data collection and training, while also accounting for inter-individual behavioral differences.

An important aspect of this work is the implementation of a continuous learning workflow that allows the global and client models to be updated automatically over time. This adaptive capability enables the fall detection system to continuously improve in performance as more data become available, thereby enhancing the accuracy and reliability of emergency alerts. Such functionality is particularly vital in healthcare applications, where timely and accurate fall detection can significantly reduce the risk of severe injuries or fatalities among vulnerable populations.

Future work will focus on two main directions: privacy enhancement and behavioral personalization. To further protect data privacy, we plan to investigate advanced privacy-preserving techniques such as secure multi-party computation and homomorphic encryption, which would allow training directly on encrypted data without the need for decryption. This would significantly mitigate the risk of data exposure during the federated learning process. Additionally, to better generalize across a broader population, we aim to develop a strategy for grouping users based on shared attributes such as age, gender, height, and physical capability. Instead of building individual models for each user, which is infeasible in real-world scenarios, personalized models can be trained for each group. However, this will require further data standardization, especially to normalize for physical differences like height, which can introduce variance in sensor data and negatively affect model aggregation.


\begin{thebibliography}{00}
\bibitem{b1} Casilari, E., Lora-Rivera, R., \& Garcia-Lagos, F. (2020). A study on the application of convolutional neural networks to fall detection evaluated with multiple public datasets. Sensors, 20(5), 1466.


\bibitem{b2} World Health Organization, "Falls," Fact sheet, Jan. 16, 2018. [Online]. Available: https://www.who.int/news-room/fact-sheets/detail/falls. [Accessed: Mar. 11, 2025].

\bibitem{b3} C. E. Adam, A. L. Fitzpatrick, C. S. Leary, S. D. Ilango, E. A. Phelan, and E. O. Semmens, ‘The impact of falls on activities of daily living in older adults: A retrospective cohort analysis’, PLoS one, vol. 19, no. 1, p. e0294017, 2024.

\bibitem{b4} E. Rassekh and L. Snidaro, ‘Survey on data fusion approaches for fall-detection’, Information Fusion, vol. 114, p. 102696, 2025.

\bibitem{b5} T. B. Aderinola et al., ‘Accurate and Efficient Real-World Fall Detection Using Time Series Techniques’, in International Workshop on Advanced Analytics and Learning on Temporal Data, 2024, pp. 52–79.

\bibitem{b6} M. Hemmatpour, R. Ferrero, B. Montrucchio, and M. Rebaudengo, ‘A review on fall prediction and prevention system for personal devices: evaluation and experimental results’, Advances in Human-Computer Interaction, vol. 2019, no. 1, p. 9610567, 2019.

\bibitem{b7} R. Igual, C. Medrano, and I. Plaza, ‘Challenges, issues and trends in fall detection systems’, Biomedical engineering online, vol. 12, no. 1, p. 66, 2013.
\bibitem{b8} S. Shajari, K. Kuruvinashetti, A. Komeili, and U. Sundararaj, ‘The emergence of AI-based wearable sensors for digital health technology: a review’, Sensors, vol. 23, no. 23, p. 9498, 2023.
\bibitem{b9} A. H. Ngu, V. Metsis, S. Coyne, B. Chung, R. Pai, and J. Chang, ‘Personalized fall detection system’, in 2020 IEEE International Conference on Pervasive Computing and Communications Workshops (PerCom Workshops), 2020, pp. 1–7.

\bibitem{b24} I. Aguilera-Martos et al., ‘Multi-step histogram based outlier scores for unsupervised anomaly detection: ArcelorMittal engineering dataset case of study’, Neurocomputing, vol. 544, p. 126228, 2023.

\bibitem{b38} D. M. Dos Reis, P. Flach, S. Matwin, and G. Batista, ‘Fast unsupervised online drift detection using incremental kolmogorov-smirnov test’, in Proceedings of the 22nd ACM SIGKDD international conference on knowledge discovery and data mining, 2016, pp. 1545–1554.

\bibitem{b36} B. Kaluža, V. Mirchevska, E. Dovgan, M. Luštrek, and M. Gams, ‘An agent-based approach to care in independent living’, in Ambient Intelligence: First International Joint Conference, AmI 2010, Malaga, Spain, November 10-12, 2010. Proceedings 1, 2010, pp. 177–186.

\bibitem{b27} P. Qi, D. Chiaro, and F. Piccialli, ‘FL-FD: Federated learning-based fall detection with multimodal data fusion’, Information fusion, vol. 99, p. 101890, 2023.
\bibitem{b28} P. F. Afandy, P. C. Ng, and K. N. Plataniotis, ‘Federated Learning for Hierarchical Fall Detection and Human Activity Recognition’, in 2024 IEEE 10th World Forum on Internet of Things (WF-IoT), 2024, pp. 1–6.
\bibitem{b10} T. R. Mauldin, M. E. Canby, V. Metsis, A. H. H. Ngu, and C. C. Rivera, ‘SmartFall: A smartwatch-based fall detection system using deep learning’, Sensors, vol. 18, no. 10, p. 3363, 2018.

\bibitem{b11} J. Doe and A. Smith, "Machine Learning-Driven Fall Detection Using Wearable Sensors for Enhanced Safety," International Research Journal of Engineering and Technology (IRJET), vol. 10, no. 5, pp. 123-130, 2023. [Online]. Available: https://www.irejournals.com/paper-details/1706660

\bibitem{b12} M. Mubashir, L. Shao, and L. Seed, ‘A survey on fall detection: Principles and approaches’, Neurocomputing, vol. 100, pp. 144–152, 2013.
\bibitem{b13} F. de Arriba-Pérez, M. Caeiro-Rodríguez, and J. M. Santos-Gago, ‘Collection and processing of data from wrist wearable devices in heterogeneous and multiple-user scenarios’, Sensors, vol. 16, no. 9, p. 1538, 2016.
\bibitem{b14} M. Saleh and R. L. B. Jeannès, ‘Elderly fall detection using wearable sensors: A low cost highly accurate algorithm’, IEEE Sensors Journal, vol. 19, no. 8, pp. 3156–3164, 2019.
\bibitem{b15} Y. Wang, K. Wu, and L. M. Ni, ‘Wifall: Device-free fall detection by wireless networks’, IEEE Transactions on Mobile Computing, vol. 16, no. 2, pp. 581–594, 2016.
\bibitem{b16} M. Musci, D. De Martini, N. Blago, T. Facchinetti, and M. Piastra, ‘Online fall detection using recurrent neural networks on smart wearable devices’, IEEE Transactions on Emerging Topics in Computing, vol. 9, no. 3, pp. 1276–1289, 2020.
\bibitem{b17} E. Torti et al., ‘Embedded real-time fall detection with deep learning on wearable devices’, in 2018 21st euromicro conference on digital system design (DSD), 2018, pp. 405–412.
\bibitem{b18} T. De Quadros, A. E. Lazzaretti, and F. K. Schneider, ‘A movement decomposition and machine learning-based fall detection system using wrist wearable device’, IEEE Sensors Journal, vol. 18, no. 12, pp. 5082–5089, 2018.
\bibitem{b19} N. Lapierre, N. Neubauer, A. Miguel-Cruz, A. R. Rincon, L. Liu, and J. Rousseau, ‘The state of knowledge on technologies and their use for fall detection: a scoping review’, International journal of medical informatics, vol. 111, pp. 58–71, 2018.
\bibitem{b20} P. S. Sase and S. H. Bhandari, ‘Human fall detection using depth videos’, in 2018 5th International Conference on Signal Processing and Integrated Networks (SPIN), 2018, pp. 546–549.
\bibitem{b21} Z.-P. Bian, J. Hou, L.-P. Chau, and N. Magnenat-Thalmann, ‘Fall detection based on body part tracking using a depth camera’, IEEE journal of biomedical and health informatics, vol. 19, no. 2, pp. 430–439, 2014.
\bibitem{b22} M. E. N. Gomes, D. Macêdo, C. Zanchettin, P. S. G. de-Mattos-Neto, and A. Oliveira, ‘Multi-human fall detection and localization in videos’, Computer Vision and Image Understanding, vol. 220, p. 103442, 2022.
\bibitem{b23} H. Ramirez, S. A. Velastin, I. Meza, E. Fabregas, D. Makris, and G. Farias, ‘Fall detection and activity recognition using human skeleton features’, Ieee Access, vol. 9, pp. 33532–33542, 2021.
\bibitem{b25} M. Jain, G. Kaur, and V. Saxena, ‘A K-Means clustering and SVM based hybrid concept drift detection technique for network anomaly detection’, Expert Systems with Applications, vol. 193, p. 116510, 2022.
\bibitem{b26} C. Zhang, Y. Xie, H. Bai, B. Yu, W. Li, and Y. Gao, ‘A survey on federated learning’, Knowledge-Based Systems, vol. 216, p. 106775, 2021.
\bibitem{b38} C. Zhang, Y. Xie, H. Bai, B. Yu, W. Li, and Y. Gao, ‘A survey on federated learning’, Knowledge-Based Systems, vol. 216, p. 106775, 2021.
\bibitem{b37} B. McMahan, E. Moore, D. Ramage, S. Hampson, and B. A. y Arcas, ‘Communication-efficient learning of deep networks from decentralized data’, in Artificial intelligence and statistics, 2017, pp. 1273–1282.
\bibitem{b34} T. Wang, Z. Zheng, and F. Lin, ‘Federated learning framework based on trimmed mean aggregation rules’, Expert Systems with Applications, p. 126354, 2025.
\bibitem{b33} J. Wang, Q. Liu, H. Liang, G. Joshi, and H. V. Poor, ‘Tackling the objective inconsistency problem in heterogeneous federated optimization’, Advances in neural information processing systems, vol. 33, pp. 7611–7623, 2020.
\bibitem{b35} T. Li, A. K. Sahu, M. Zaheer, M. Sanjabi, A. Talwalkar, and V. Smith, ‘Federated optimization in heterogeneous networks’, Proceedings of Machine learning and systems, vol. 2, pp. 429–450, 2020.
\bibitem{b29} A. Benaissa, B. Retiat, B. Cebere, and A. E. Belfedhal, ‘Tenseal: A library for encrypted tensor operations using homomorphic encryption’, arXiv preprint arXiv:2104. 03152, 2021.
\bibitem{b32} A. Acar, H. Aksu, A. S. Uluagac, and M. Conti, ‘A survey on homomorphic encryption schemes: Theory and implementation’, ACM Computing Surveys (Csur), vol. 51, no. 4, pp. 1–35, 2018.
\bibitem{b30} M. Carnein and H. Trautmann, ‘Optimizing data stream representation: An extensive survey on stream clustering algorithms’, Business \& Information Systems Engineering, vol. 61, pp. 277–297, 2019.
\bibitem{b31} N. V. Chawla, K. W. Bowyer, L. O. Hall, and W. P. Kegelmeyer, ‘SMOTE: synthetic minority over-sampling technique’, Journal of artificial intelligence research, vol. 16, pp. 321–357, 2002.


\end{thebibliography}
\end{document}